\documentclass{article}
\usepackage{ijcai24}

\usepackage{times}
\usepackage{natbib}
\usepackage{soul}
\usepackage{url}
\usepackage[hidelinks]{hyperref}
\usepackage[utf8]{inputenc}
\usepackage[small]{caption}
\usepackage{graphicx}
\usepackage{amsmath}
\usepackage{amsthm}
\usepackage{amssymb}
\usepackage{pifont}
\usepackage{mathtools}
\usepackage{booktabs}
\usepackage{algorithm}
\usepackage{algorithmic}
\usepackage[switch]{lineno}
\usepackage{wrapfig}
\usepackage{tabularx}
\usepackage{multicol,multirow}
\usepackage[table]{xcolor}
\usepackage{colortbl}
\usepackage{makecell}
\usepackage{paralist}
\usepackage{subcaption}

\usepackage[hyperfirst=false]{glossaries}
\robustify{\Gls}
\robustify{\gls}
\newacronym{ml}{ML}{machine learning}
\newacronym{ai}{AI}{Artificial Intelligence}
\newacronym{cv}{CV}{Computer Vision}
\newacronym{nlp}{NLP}{Natural Language Processing}

\newacronym{da}{DA}{Data Augmentation}
\newacronym{cda}{CDA}{Causal Data Augmentation}
\newacronym{cdg}{CDG}{Causal Data Generation}
\newacronym{dag-cdg}{DAG-CDG}{DAG-constrained Causal Data Generation}
\newacronym{ood}{OOD}{Out-of-Domain}

\newacronym{cm}{CM}{Causal Model}
\newacronym{scm}{SCM}{structural causal model}
\newacronym{cgm}{CGM}{Causal Graphical Model}
\newacronym{cg}{CG}{Causal Graph}
\newacronym{dag}{DAG}{Directed Acyclic Graph}
\newacronym{admg}{ADMG}{Acyclic-Directed Mixed Graph}
\newacronym{mag}{MAG}{Maximal Ancestral Graph}
\newacronym{ate}{ATE}{Average Treatment Effect}

\newacronym{dscm}{DSCM}{deep structural causal model}
\newacronym{ncm}{NCM}{Neural Causal Model}
\newacronym{bgm}{BGM}{bijective generation mechanism}
\newacronym{anm}{ANM}{additive noise model}
\newacronym{lsnm}{LSNM}{location scale noise model}
\newacronym{pnl}{PNL}{post-nonlinear noise model}
\newacronym{dgm}{DGM}{deep generative model}

\newacronym{gan}{GAN}{Generative Adversarial Network}
\newacronym{vgae}{VGAE}{Variational Graph Auto-Encoders}
\newacronym{vae}{VAE}{Variational Auto-Encoders}
\newacronym{nf}{NF}{Normalizing Flow}
\newacronym{bn}{BN}{Bayesian Network}
\newacronym{kde}{KDE}{Kernel Density Estimate}
\newacronym{cnn}{CNN}{Convolutional Neural Network}
\newacronym{diff}{Diff}{Diffusion model}
\newacronym{mlp}{MLP}{Multi-Layer Perceptron}
\newacronym{fnn}{FNN}{Feedforward Neural Network}

\newacronym{gmm}{GMM}{Gaussian Mixture Model}
\newacronym{xgb}{XGBoost}{eXtreme Gradient Boosting}
\newacronym{mcts}{MCTS}{Monte Carlo Tree Search}

\newacronym{mape}{MAPE}{Mean Absolute Percentage Error}
\newacronym{mdl}{MDF}{Minimum Description Length}
\newacronym{tpr}{TPR}{True Positive Rate}
\newacronym{tnr}{TNR}{True Negative Rate}
\newacronym{shd}{SHD}{Structural Hamming Distance}
\newacronym{sid}{SID}{Structural Intervention Distance}

\newacronym{bmdc}{BMDC}{Bone Marrow-derived Dendritic Cell}

\newacronym{pc}{PC}{Peter-Clark}
\newacronym{fci}{FCI}{Fast Causal Inference}
\newacronym{ges}{GES}{Greedy Equivalence Search}

\newacronym{pch}{PCH}{Pearl Causal Hierarchy}
\newacronym{kl}{KL}{Kullback-Leibler}

\newacronym{rct}{RCT}{Randomized Control Trials}
\newacronym{cate}{CATE}{conditional average treatment effect}

\glsdisablehyper

\usepackage[capitalise]{cleveref}

\newcommand{\PA}{\textit{\textbf{PA}}}
\newcommand{\doop}[1]{\textit{\textbf{do}}(#1)}

\newcommand{\term}[1]{\textbf{#1}}
\newcommand{\ie}{i.e., }
\newcommand{\eg}{e.g., }

\newcolumntype{C}{>{\centering\arraybackslash}X}
\newcommand{\xmark}{\ding{55}}%

\title{Learning Structural Causal Models through Deep Generative Models: Methods, Guarantees, and Challenges}

\author{
Audrey Poinsot$^{1,2}$\and%
Alessandro Leite$^2$\and%
Nicolas Chesneau$^1$\and%
Mich\`ele S\'ebag$^2$\And%
Marc Schoenauer$^2$\\%
\affiliations
$^1$Ekimetrics, France\\
$^2$TAU, LISN, INRIA Saclay, France\\
\emails
audrey.poinsot@ekimetrics.com, 
alessandro.leite@inria.fr
}

\begin{document}

\maketitle

\begin{abstract}
    This paper provides a comprehensive review of~\glspl{dscm}, particularly focusing on their ability to answer counterfactual queries using observational data within known causal structures. It delves into the characteristics of~\glspl{dscm} by analyzing the hypotheses, guarantees, and applications inherent to the underlying deep learning components and structural causal models, fostering a finer understanding of their capabilities and limitations in addressing different counterfactual queries. Furthermore, it highlights the challenges and open questions in the field of deep structural causal modeling. It sets the stages for researchers to identify future work directions and for practitioners to get an overview in order to find out the most appropriate methods for their needs\@.
\end{abstract}

\section{Introduction}\label{sec:intro}
The extraction of general and scientific knowledge from extensive and complex datasets has increasingly become a critical expectation in various domains. As highlighted by~\cite{pearl:09}, attaining a deeper understanding that transcends mere associations necessitates an exploration of causation. This is particularly true when seeking to answer ``why'' questions, which require an insight into the underlying data generating processes. Standard generative approaches typically model the joint density of a set of variables. In contrast, causal generative models emphasize distributions derived from causal interventions and counterfactual operations as classified by the~\gls{pch}~\citep{pearl:18,bareinboim:22}. The~\gls{pch} comprises three cognitive layers, named association, intervention, and counterfactual, correlating with the cognitive processes of seeing, doing, and imagining~\citep{pearl:18}. A comprehensive understanding of these layers is imperative for addressing diverse causal queries. The first layer, $\mathcal{L}_1$, focuses on associations and factual data. The second one, $\mathcal{L}_2$, delves into intervention data, shedding light on the effects of specific manipulations~(a.k.a interventions). The third and final layer, $\mathcal{L}_3$, encompasses counterfactuals, facilitating the exploration of hypothetical scenarios that might have ensued under different interventions, as opposed to actual occurrences. Causal inference tasks aims to derive a quantitative understanding at a higher-level hierarchy when provided with data solely from lower-level ones~\citep{pearl:18}. 
However, achieving such a goal is inherently challenging without additional assumptions, primarily because the indeterminacy of higher levels by the lower ones poses a fundamental obstacle to such inference efforts~\citep{ibeling:20,bareinboim:22}. The so-called indeterminacy means that without making explicit assumptions about the higher level, it is impossible to answer higher-level questions from lower-level data. For instance, answering counterfactual questions requires one to model the influence of the unobserved variables~\citep{bareinboim:22}. \Glspl{scm} enable one to do so (\Cref{sec:preliminaries}).
An~\gls{scm} \(\mathcal{M}\) models a set of causal mechanisms \(\mathcal{F}\) on the observed \(\mathbf{X}\) and unobserved \(\mathbf{U}\) variables in addition to distributions over the unobserved variables \(P(\mathbf{U})\). Nevertheless, a complete \gls{scm} is rarely known in practice, 
which leads one to question how a counterfactual query~(\(\mathcal{L}_3\)) can be evaluated using a set of data from the underlying levels~(\ie{} \(\mathcal{L}_1\) and \(\mathcal{L}_2\)). In general, experimental data~(\(\mathcal{L}_2\)) are difficult to access. Thus, in practice, one relies on observational data (\(\mathcal{L}_1\)) and the knowledge of the causal structure~(\(\mathcal{L}_2\)), formulated as a causal graph~\citep{pearl:95,spirtes:00}\@.\\
\indent{} The recent advancements in deep learning, particularly the development of~\glspl{dgm}, have spawned a multitude of~\glspl{scm}, the so-called~\glspl{dscm}~\citep{pawlowski20}, which rely on deep learning approaches to model causal mechanisms. \Glspl{dgm} enable one to build more expressive~\glspl{scm}~\citep{xia2022causalneural} thanks to the universal approximation capabilities of neural networks~\citep{HORNIK1991251}. As a result, they have been used in various high-stakes domains, including healthcare and fairness, where identifiability guarantees~(\ie{} existence and uniqueness) of the sought model are fundamental. However, it is essential to acknowledge that different~\glspl{dscm} may operate under diverse assumptions regarding the functional form of causal mechanisms, leading to distinct guarantees. Understanding these variations is pivotal for informed and effective application of~\glspl{dscm} in different contexts.\\ 
\indent{} This paper aims to clarify the scope and limitations of the existing~\glspl{dscm} when provided a known causal graph to answer counterfactual queries. Our primary goal is to make a tour of the state-of-the-art, enabling a practitioner to better identify their needs and the requisites to fulfill them. A secondary one is to list the open questions and challenges. 
Formally, the overarching question investigated by this paper can be articulated as follows. Given a known causal structure \(\mathcal{G}^\star\) and observational data, what are the capabilities of existing~\glspl{dgm} in learning a~\gls{scm} \(\widehat{\mathcal{M}}\) that approximates the true but unknown \gls{scm} \(\mathcal{M}^\star\)? In other words, we are interested in their capability in answering counterfactual questions as if one had access to the true~\gls{scm} \(\mathcal{M}^\star\). 
\paragraph{Contributions.} The contributions of this paper are threefold. First, it categorizes the existing~\acrlongpl{dscm}~(\Cref{sec:taxonomy}) following their assumptions and guarantees based solely on observational data and a known causal structure. This classification provides a comprehensive framework for understanding the landscape of deep structural causal modeling when aiming to answer counterfactual questions. Second, analyzing existing~\gls{dscm} from a theoretical and a empirical viewpoint, this paper assists practitioners when selecting an appropriate~\gls{dgm} to learn an~\gls{scm}~(\Cref{sec:model_selection}) in order to apply these models effectively. Finally, this paper highlights some open questions and challenges~(\Cref{sec:challenges}), serving as suggestions for future research directions in the field. Particularly, 
this paper stands out as the first one to classify the methods to learn~\glspl{scm} through~\gls{dgm} based on their hypotheses and guarantees. In contrast, existing surveys~\citep{zhou2023emerging, komanduri:survey:23, kaddour2022causal} primarily classify them according to their deep learning components\@.
%
\section{Preliminaries of Causal Inference}\label{sec:preliminaries}

In this work, we use capital letters for random variables, lowercase letters for their realizations, and bold letters to represent vectors.

\subsection{Structural Causal Model}\label{sec:scm}

A~\term{Structural Causal Model}~\citep{pearl:09} is a tuple \(\mathcal{M} \coloneqq (\mathcal{F}, P(\boldsymbol{U}))\), where \(\mathcal{F}\) comprises a set of $d$ structural equations \(f_i\), one for each random variable, \(X_i \in \boldsymbol{X}\), \(\mathcal{F} = \{X_i \coloneqq f_i(\PA(X_i), U_i)\}^{d}_{i=1}\). Each structural equation \(f_i\) computes each variable \(X_i\), also called endogenous variable, from its parents \(\PA(X_i) \subseteq \boldsymbol{X} \setminus \{X_i\}\), and an exogenous noise variable \(U_i \in \boldsymbol{U}\). \(P(\mathbf{U})\) is a strictly positive measure over the exogenous noises, \(\boldsymbol{U}\). Whenever an unmeasured common cause exists for several endogenous variables, these variables are said to be confounded by a hidden confounder. If one assumes that no hidden confounder exists over the endogenous variables $\mathbf{X}$, the exogenous noise variables $\mathbf{U}$ are considered jointly independent. An~\gls{scm} characterizes a unique distribution over the variables, \(P_{\mathcal{M}}(\boldsymbol{X})\), called the entailed distribution. 
The conditional distribution of a variable given its parents, $P_{\mathcal{M}}(X_i|\PA(X_i))$, is called a causal kernel~\citep{pearl:09}\@. %
An~\gls{scm} describes the so-called~\term{causal graph} \(\mathcal{G} \coloneqq (\mathcal{V}, \mathcal{E})\) which encodes the causal dependencies between the variables. Each node corresponds to an endogenous variable, and the set of directed edges represents the parent-child relationships; \ie{} $\mathcal{V}=\boldsymbol{X}$ and $\mathcal{E} = \{\PA(X_i) \xrightarrow{} X_i, \, \forall i \in [1,d]\}$. 
A causal graph is usually assumed to be a~\gls{dag}.\\
\indent
One can use an~\gls{scm} to answer intervention and counterfactual questions~\citep{pearl:09survey}. 
Given an~\gls{scm} $\mathcal{M}$, an~\term{intervention} consists in changing at least one of the structural equations, for instance, $X_k \coloneqq \widetilde{f}_k(\widetilde{\PA}(X_k), \widetilde{U}_k)$. This operation, denoted $\doop{X_k \coloneqq \widetilde{f}_k(\widetilde{\PA}(X_k), \widetilde{U}_k)}$, defines a new~\gls{scm} $\widetilde{\mathcal{M}}$ whose entailed distribution is called the intervention distribution, $P_{\widetilde{\mathcal{M}}}(\boldsymbol{X}) = P_{\mathcal{M}}(\boldsymbol{X}|\doop{X_k \coloneqq \widetilde{f}_k(\widetilde{\PA}(X_k), \widetilde{U}_k)})$. 
Given an~\gls{scm} $\mathcal{M}$ and some factual realizations $\boldsymbol{y}$ of $\boldsymbol{Y} \subset \boldsymbol{X}$, a~\term{counterfactual} consists in changing the exogenous noise distribution into the posterior distribution $P_{\boldsymbol{U}|\boldsymbol{Y}=\boldsymbol{y}}$. It defines a new~\Gls{scm} $\mathcal{M}^{\prime}$, where $P_{\mathcal{M}^{\prime}}(\boldsymbol{U}) = P_{\mathcal{M}}(\boldsymbol{U}|\boldsymbol{Y}=\boldsymbol{y})$, whose entailed distribution is called the counterfactual distribution. Counterfactuals generally also include an additional intervention. The counterfactual distribution is hence denoted by $P_{\mathcal{M}}^{\boldsymbol{Y}=\boldsymbol{y}}(\boldsymbol{X}|\doop{X_k \coloneqq \widetilde{f}_k(\widetilde{\PA}(X_k), \widetilde{U}_k)})$. Counterfactuals can be derived through the following three-step procedure~\citep{pearl:09}: (a) \textit{Abduction} to change the exogenous noise distribution, (b) \textit{Action} to build the new~\gls{scm} $\mathcal{M}^{\prime}$, and (c) \textit{Prediction} to estimate the counterfactual distribution given $\mathcal{M}^{\prime}$.

\subsection{Identifiability}\label{sec:identifiabiliy}

Once a causal task is defined in terms of its causal quantities, it needs to be translated into statistical ones for estimation. The property of identification amounts to showing whether, under the appropriate set of assumptions, such a unique translation exists.
%
There exist several types of identifiability. For instance, structure identifiability states whether the causal graph can be recovered, model identifiability whether the full causal model can be recovered, and, estimand identifiability whether a specific causal estimand, such as a treatment effect, can be estimated. Finally, \term{intervention and counterfactual identifiability} involves checking whether all intervention and counterfactual queries can be retrieved. Formally, given a class of causal models $\mathbb{M}$ (\eg{} \glspl{scm}), an $\mathcal{L}_i$-query $Q(\mathcal{M})$ of a model $\mathcal{M} \in \mathbb{M}$ is identifiable from $\mathcal{L}_j$-data, $1\leq j<i$, if for any pair of models $\mathcal{M}_1$ and $\mathcal{M}_2$ from $\mathbb{M}$, $Q(\mathcal{M}_1)=Q(\mathcal{M}_2)$ whenever $\mathcal{M}_1$ and $\mathcal{M}_2$ match in all $\mathcal{L}_j$ queries~\citep{pearl:09}. Let us note that model identifiability implies intervention and counterfactual identifiability.

\section{Classification into SCM and DGM Classes}\label{sec:taxonomy}

\begin{figure}
  \centering
  \includegraphics[trim=220 270 260 72,clip, width=.99\columnwidth]{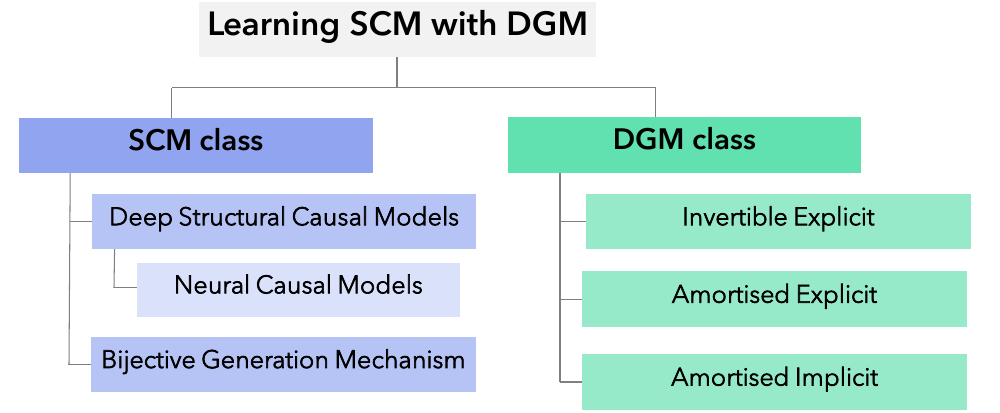}
  \caption{The two-dimension classification of~\glspl{dscm}}\label{fig:taxonomy}
\end{figure}

This section introduces a classification of~\glspl{dgm} designed to learn an approximate~\acrlong{scm} \(\widehat{\mathcal{M}}\) when provided a known causal graph \(\mathcal{G}^\star\) and observational data. By abuse of language but without loss of generality, we refer to these methods as~\glspl{dscm}, since they all model causal mechanisms with deep generative models. The purpose of the approximate~\acrlong{scm} \(\widehat{\mathcal{M}}\) is to answer counterfactual queries. In practical terms, a method aiming to answer counterfactual queries must effectively perform the abduction step by modeling the distribution of the exogenous variables given the realization of the factual data~(\ie{} \(P(\boldsymbol{U}\mid{} \mathbf{X} = \mathbf{x}\)) by using the causal mechanisms (\Cref{sec:scm}). In addition, assessing whether $\mathcal{L}_3$-identifiability holds is paramount to build trust in the estimated counterfactuals.\\
\indent The classification, illustrated in~\cref{fig:taxonomy}, comprises two dimensions. The first dimension relates to the classes of~\gls{scm} providing identifiability guarantees~(\Cref{tb:scm_class_hypotheses}), while the second one, originally proposed by~\cite{pawlowski20}, focuses on the characteristics of the~\gls{dgm} used to learn the~\gls{scm}, restricting hence the tractability of the abduction step~(\Cref{tb:dl_approach_abduction}). In this classification, \gls{dgm} classes are exclusive while the~\gls{scm} ones are not, as one can see in~\cref{tb:model_hypotheses}.

\begin{table*}
    \begin{subtable}{.46\linewidth}
      \centering
        \footnotesize{%
        \begin{tabularx}{\linewidth}{>{\hsize=.11\hsize}C>{\hsize=.3\hsize}X>{\hsize=.59\hsize}X}
          \toprule
            \multicolumn{1}{c}{\textbf{Class}}  & \multicolumn{1}{c}{\textbf{Mechanism}} & \multicolumn{1}{c}{\textbf{Identifiability Guarantees}} \\
          \midrule
    
            \textsc{\textit{DSCM}} &  
            Neural network & 
            - \\
    
            
    

            \rowcolor{gray!25}
            \textsc{\textit{NCM}} &  
            \makecell[l]{Feedforward \\ neural network} & 
            \makecell[l]{$\mathcal{L}_3$-id. (resp. $\mathcal{L}_2$) iif $\mathcal{L}_3$-id. \\ (resp. $\mathcal{L}_2$) from the true~\gls{scm}} \\

            \textsc{\textit{BGM}} &  
            Bijective noise & 
            \makecell[l]{$\mathcal{L}_3$-id. for three settings\(^\star\) cf. \\ Theorems 5.1, 5.2 and 5.3}\\
        \cmidrule{1-3}
        \multicolumn{3}{l}{\makecell[l]{\(^\star\)Markovian, Instrumental Variable, and Backdoor Criterion}}\\
       \bottomrule
     \end{tabularx}
     \caption{Identifiability guarantees of the classes of~\glspl{scm}}\label{tb:scm_class_hypotheses}
    }
    \end{subtable}%
    \hspace{11pt}
    \begin{subtable}{0.5\linewidth}
        \centering
        \footnotesize{%
        \begin{tabularx}{\linewidth}{>{\hsize=.37\hsize}C>{\hsize=.21\hsize}C>{\hsize=.21\hsize}C>{\hsize=.21\hsize}C}
          \toprule
            & \multicolumn{3}{c}{\textbf{Abduction step}}\\
          \cmidrule(lr){2-4}
            \multirow{1}{*}{\textbf{\makecell[c]{Class of~\gls{dgm}}}}  & Mechanism Inversion & Encoding & Sample Rejection \\
          \midrule
    
            \textsc{\textit{Invertible Explicit}} &  
            \checkmark & 
            \checkmark & 
            \checkmark \\
            
            \rowcolor{gray!25}
            \textsc{\textit{Amortised Explicit}} &  
            \xmark & 
            \checkmark & 
            \checkmark \\
            
            \textsc{\textit{Amortised Implicit}} &  
            \xmark & 
            \xmark & 
            \checkmark \\
            
       \bottomrule
     \end{tabularx}
     \caption{Abduction steps for the classes of~\glspl{dgm}}\label{tb:dl_approach_abduction}
    }
    \end{subtable}%
    \caption{Identifiability and abduction step guarantees of SCM and DGM classes}\label{tb:taxonomy_classes}
\end{table*}

\subsection{Classes of Deep Generative Models}\label{sec:dscm}

\Glspl{dscm} rely on deep learning components to capture and model the causal mechanisms. The choice of the network architecture can make it hard to compute the abduction step. Hence, a~\gls{dscm} falls into one of the following classes introduced by~\cite{pawlowski20}: invertible explicit, amortised explicit, and amortised implicit.

\paragraph{Invertible Explicit.}
Invertible explicit methods use invertible mechanisms with regard to the exogenous noises. The abduction step is hence directly tractable, computing the inverse transformation of the mechanism. Existing invertible explicit methods use conditional normalizing flows to represent the structural equations. In this case, each causal kernel can be computed as $P(X_i | \PA(X_i))=P(U_i) \cdot |det\nabla_{U_i}f_i(\PA(X_i), U_i)|^{-1}|_{U_i=f_i^{-1}(U_i, \PA(X_i))}$ whenever $f_i$ is assumed to be a diffeomorphic transformation. Examples of methods include \textbf{Causal-NF}~\citep{javaloy2023causal}, \textbf{NF-DSCM}~\citep{pawlowski20}, \textbf{NCF}~\citep{parafita2020causal}
, \textbf{CAREFL}~\citep{khemakhem21a} and, \textbf{NF-BGM}~\citep{nasr_23}. 

\paragraph{Amortised Explicit.}
Amortised explicit methods enlarge the considered causal mechanisms to non-invertible functions.
By losing the invertibility of the structural equations, the abduction step can no longer be directly computed. Hence, these methods approximate the conditional exogenous noise distribution through auto-encoders. Notably, $f_i(\PA(X_i), U_i) = g_i(\PA(X_i), U_i)$ and $U_i = e_i(\PA(X_i), X_i)$ where $g_i$ and $e_i$ represent the decoder and the encoder\footnote{Note that the initial functional form enables ``low-level'' transformations: $f_i(\PA(X_i), U_i) = h_i(\PA(X_i), g_i(\PA(X_i),\mu_i),\epsilon_i)$ where $g_i$ is a function of any form, $h_i$ an invertible function, and $U_i = (\epsilon_i, \mu_i)$ a noise decomposition such that $P(U_i) = P(\epsilon_i)P(\mu_i)$.}. Examples of learning methods belonging to this class include \textbf{iVGAE}~\citep{zecevic2021relating}, \textbf{VACA}~\citep{sanchezmartin:22}, and, \textbf{DCM}~\citep{chao2023interventional}.

\paragraph{Amortised Implicit.}
Instead of explicitly learning an encoder, one can also implicitly model it, for instance, using adversarial learning. Such approaches constitute the amortised implicit class. In this case, the abduction step must be conducted using sample rejection~\citep{xia2023neural}. In other words, for a given observation $\boldsymbol{Y}=\boldsymbol{y}$ and any~\gls{scm}, one needs to sample and filter the exogenous noises not generating the desired observation $\boldsymbol{y}$; and then, uses the remained samples to estimate the counterfactual distribution corresponding to $P(U|\boldsymbol{Y}=\boldsymbol{y})$. This class includes the following learning methods: \textbf{SCM-VAE}~\citep{Komanduri22}, \textbf{Causal-TGAN}~\citep{wen2022causaltgan}, \textbf{CausalGAN}~\citep{kocaoglu2017causalgan}, \textbf{CFGAN}~\citep{ijcai2019p201}, \textbf{DECAF}~\citep{vanBreugel21}, \textbf{WhatIfGAN}~\citep{rahman2023towards}, \textbf{CGN}~\citep{sauer2021counterfactual}, \textbf{DEAR}~\citep{shen_22}, \textbf{GAN-NCM} and \textbf{MLE-NCM}~\citep{xia2023neural}.
\subsection{Classes of Structural Causal Models}
As stated in the previous section, there are various deep generative approaches that can be used to learn an~\gls{scm}. They all model~\glspl{dscm} as they rely on deep learning components. However, they make different assumptions about the causal mechanisms leading to differences in the modelled~\glspl{dscm}. In particular, the Bijective Generation Mechanisms~\citep{nasr_23} and Neural Causal Models~\citep{xia2022causalneural} are two classes of~\glspl{scm} to distinguish them.
\subsubsection{Bijective Generation Mechanisms}\label{sec:bgm}
\Gls{bgm} models~\citep{nasr_23} are~\acrlongpl{scm} where the causal mechanisms $f_i:\{\PA(X_i), U_i\} \mapsto X_i$ are bijective mappings from the exogenous noises to the endogenous variables. \Gls{bgm} models can also represent hidden confounders: if a hidden confounder has an impact on both $X_i$ and $X_j$, their exogenous noises $U_i$ and $U_j$ are considered dependent. The bijective mapping enables to uniquely identify the exogenous noise $U_i$ for each realization of the direct causes $\PA(X_i)$. 
As a result, all the invertible explicit methods are notably modeling~\gls{bgm} models. 
\cite{nasr_23} provides sets of constraints on the causal mechanisms and probability distributions to guarantee $\mathcal{L}_3$-identifiability for three common causal structures: bivariate Markovian, Instrumental Variable, and Backdoor Criterion\@.

\subsubsection{Neural Causal Models}\label{sec:ncm}
The class of~\glspl{ncm} is introduced by~\citep{xia2022causalneural}. It defines a set of~\glspl{dscm} whose causal mechanisms are feedforward neural networks. They involve all generative models except diffusion models (\eg{} GANs, VAE, NF). Unlike feedforward neural networks, which process data in a direct, layer-by-layer flow without feedback, diffusion models employ an iterative and inherently stochastic mechanism. It involves gradually adding noise to the data over a series of steps to form a noisy distribution, and then systematically reversing this process through feedback mechanisms to generate new samples from the noisy distribution~\citep{sohl:15deep}. As a result, except DCM, all the methods mentioned previously model~\glspl{ncm}.~\Glspl{ncm} are defined to consider hidden confounders. Besides the former exogenous noise (\ie{} one per observed variable),~\glspl{ncm} consider additional exogenous noises, one for each hidden confounder impacting a subset $C$ of variables: $X_i=f_i(PA(X_i), \{U_C | X_i \in C\})$. The distributions of the exogenous variables are set to $Unif(0,1)$\footnote{Note that, as the exogenous noise variable enters as input of the neural model, the assumption of uniform noise is not limiting.}. It is proven that an $\mathcal{L}_2$~\citep{xia2022causalneural} or $\mathcal{L}_3$~\citep{xia2023neural} query is identifiable from an~\gls{ncm} if and only if it is identifiable from the true~\gls{scm}. Indeed, given that there might exist some hidden confounders, neither $\mathcal{L}_2$ nor $\mathcal{L}_3$ identifiability are guaranteed in general.

\section{Analysis of Deep Structural Causal Models}\label{sec:model_selection}
This section analyses the existing \glspl{dscm} under the hypotheses, guarantees, evaluation, and applications. \Cref{sec:hyp_guarantees} describes the hypotheses and guarantees under five distinct characteristics~(\Cref{tb:model_hypotheses}): \begin{inparaenum}[(a)] \item \textbf{causal structure knowledge}: the required type of knowledge on the causal structure, \item \textbf{hidden confounding}: whether the method can handle hidden confounding, \item \textbf{data assumption}: the hypotheses on the causal generating process of the data, \item \textbf{abduction}: the capacity of the method to perform the abduction step, and, \item the \textbf{additional guarantees} highlight supplemental guarantees to the ones provided by the \gls{scm} class\end{inparaenum}. \Cref{sec:perf} focuses on the practical capabilities of the existing methods based on the experiments reported in the corresponding papers and the existing applications~(\Cref{tb:model_evaluation_nb}).

\subsection{Hypotheses and Guarantees}\label{sec:hyp_guarantees}


\defcitealias{nasr_23}{[1]}
\defcitealias{pawlowski20}{[2]}
\defcitealias{xia2023neural}{[3]}
\defcitealias{javaloy2023causal}{[4]}
\defcitealias{parafita2020causal}{[5]}
\defcitealias{khemakhem21a}{[6]}
\defcitealias{zecevic2021relating}{[7]}
\defcitealias{sanchezmartin:22}{[8]}
\defcitealias{chao2023interventional}{[9]}
\defcitealias{Komanduri22}{[10]}
\defcitealias{wen2022causaltgan}{[11]}
\defcitealias{kocaoglu2017causalgan}{[12]}
\defcitealias{ijcai2019p201}{[13]}
\defcitealias{vanBreugel21}{[14]}
\defcitealias{rahman2023towards}{[15]}
\defcitealias{sauer2021counterfactual}{[16]}
\defcitealias{shen_22}{[17]}

\begin{table*}[htb]
    \centering
    \footnotesize{%
    \begin{tabularx}{\linewidth}{>{\hsize=.16\hsize}C>{\hsize=.12\hsize}C>{\hsize=.05\hsize}C>{\hsize=.07\hsize}C>{\hsize=.09\hsize}C>{\hsize=.17\hsize}X>{\hsize=.09\hsize}C>{\hsize=.27\hsize}X}
      \toprule
        & \multicolumn{2}{c}{\textbf{Classification}} & \multicolumn{4}{c}{\textbf{Additional Hypotheses}} & \multicolumn{1}{c}{\textbf{Additional Guarantees}}\\
      \cmidrule(lr){2-3} \cmidrule(lr){4-7} \cmidrule(lr){8-8}
        \multirow{1}{*}{\textbf{Method}} & SCM class & DGM class\(^\star\) & \makecell[l]{Causal \\ Structure} & \makecell[l]{Hidden \\ Confounder} & Data Assumptions & \makecell[l]{Available \\ Abduction} & \makecell[l]{Identifiability, Expressivity, \\ Bounds} \\
      \midrule

        \rowcolor{gray!25}
        \textsc{\textit{NF-BGM}} \citetalias{nasr_23} &  
        BGM, NCM &
        IE &
        DAG & 
        \checkmark & 
        - & 
        Inversion & 
        - \\
        
        


        \textsc{\textit{NF-DSCM}} \citetalias{pawlowski20} &  
        BGM, NCM &
        IE &
        DAG & 
        \xmark & 
        $f_i$ diffeomorphic & 
        Inversion & 
        - \\

        \rowcolor{gray!25}
        \textsc{\textit{GAN-NCM; MLE-NCM}} \citetalias{xia2023neural} &  
        NCM &
        AI &
        DAG & 
        \checkmark\(^\sharp\) & 
        - & 
        Sample Rejection & 
        - \\

        \textsc{\textit{Causal-NF}} \citetalias{javaloy2023causal} &  
        BGM, NCM &
        IE &
        Ordering & 
        \xmark & 
        $f_i$ diffeomorphic & 
        Inversion & 
        Model id. up to invertible transformation of $U$ \\
        
        \rowcolor{gray!25}
       \textsc{\textit{NCF}} \citetalias{parafita2020causal} &  
        BGM, NCM &
        IE &
        DAG & 
        \checkmark \(^\sharp\) & 
        $f_i$ diffeomorphic & 
        Inversion & 
        - \\

        \textsc{\textit{CARFEL}} \citetalias{khemakhem21a} &  
        BGM, NCM &
        IE &
        DAG, $\emptyset$ & 
        \xmark & 
        Affine autoregressive flow & 
        Inversion & 
        Model id. in bivariate case\\
        
        \rowcolor{gray!25}
        \textsc{\textit{iVGAE}} \citetalias{zecevic2021relating} &  
        NCM &
        AE &
        DAG & 
        \xmark & 
        - & 
        \xmark & 
        - \\

        \textsc{\textit{VACA}} \citetalias{sanchezmartin:22} &  
        NCM &
        AE &
        DAG & 
        \xmark & 
        - & 
        Encoding & 
        $\mathcal{L}_2$-expressivity if the decoder is deep enough cf. Prop.2 \\
        
        \rowcolor{gray!25}
        \textsc{\textit{DCM}} \citetalias{chao2023interventional} &   
        - &
        AE &
        Ordering & 
        \xmark & 
        - & 
        Encoding & 
        $\mathcal{L}_3$-id. with error bounds cf. Corollary 1 \& 2 \\

        \textsc{\textit{SCM-VAE}} \citetalias{Komanduri22} &  
        NCM &
        AI &
        DAG & 
        \xmark & 
        Additive noise on attributes & 
        \xmark & 
        - \\
        
        \rowcolor{gray!25}
        \textsc{\textit{Causal-TGAN}} \citetalias{wen2022causaltgan} &   
        NCM &
        AI &
        DAG & 
        \xmark & 
        - & 
        \xmark & 
        - \\

        \textsc{\textit{CausalGAN}} \citetalias{kocaoglu2017causalgan} &   
        NCM &
        AI &
        DAG & 
        \xmark & 
        - & 
        \xmark & 
        - \\
        
        \rowcolor{gray!25}
        \textsc{\textit{CFGAN}} \citetalias{ijcai2019p201} &   
        NCM &
        AI &
        DAG & 
        \xmark & 
        Categ. outcome \& sensitive feature &
        \xmark & 
        - \\

        \textsc{\textit{DECAF}} \citetalias{vanBreugel21} &   
        NCM &
        AI &
        DAG & 
        \xmark & 
        - & 
        \xmark & 
        - \\
        
        \rowcolor{gray!25}
        \textsc{\textit{WhatIfGAN}} \citetalias{rahman2023towards} &   
        NCM &
        AI &
        DAG & 
        \checkmark & 
        - & 
        \xmark & 
        - \\

        \textsc{\textit{CGN}} \citetalias{sauer2021counterfactual} &   
        NCM &
        AI &
        DAG\(^\tau\) & 
        \checkmark \(^\tau\) & 
        Image with attributes & 
        \xmark & 
        - \\
        
        \rowcolor{gray!25}
        \textsc{\textit{DEAR}} \citetalias{shen_22} &   
        NCM &
        AI &
        Ordering & 
        \xmark & 
        High-dimensional data with attributes & 
        \xmark & 
        Data to attribute encoder disentanglement \\
    \cmidrule{1-8}
    \multicolumn{8}{l}{\(^\sharp\)A common cause is represented by an additional exogenous noise, \(^\tau\)Only a confounded trivariate~\gls{dag} is considered}\\
    \multicolumn{8}{l}{\(^\star\)Invertible Explicit (IE), Amortised Explicit (AE), and Amortised Implicit~(AI)}\\
   \bottomrule
 \end{tabularx}
}
\caption{Hypotheses and guarantees of deep structural causal models. The classification (\Cref{fig:taxonomy}) enables one to spot the identifiability results inherited by the~\gls{scm} class and the compatible abduction step procedures.}\label{tb:model_hypotheses}
\end{table*}

\paragraph{Causal Structure Knowledge.} Some approaches only need a causal ordering as input. Indeed, a causal ordering is usually sufficient to build a fully connected~\gls{dag} compatible, but non-minimal, with the true~\gls{scm}. However, this is at the expense of additional training, computation time, and complexity. Other methods rely on the known causal graph.
Note that CARFEL can also learn the causal structure on top of the causal mechanisms. However, in this case, model identification is only provided in the Markovian bivariate setting. 

\paragraph{Hidden confounding.} As said, unobserved common causes create spurious correlations that bias $\mathcal{L}_3$ estimations, and real-world applications tend to involve hidden confounding. However, only five methods can deal with them. CGN considers confounded variables in a specific structure configuration, while NCM, BGM, WhatIfGAN, and NCF can deal with any position of unobserved variables in the causal graph as long as this position is known. NeuralID~\citep{xia2023neural}, an algorithm built on top of the NCM class, can automatically evaluate if a query is identifiable given a~\gls{dag} with hidden confounders. To do so, two NCMs, constrained to preserve consistency with the observed data, are trained in parallel to maximize and minimize the target query. If the two resulting queries are identical, then the query is considered identifiable. Such a tool is crucial for sensitive applications.

\paragraph{Data Assumptions.} Apart from the methods dealing with images (\ie{} SCM-VAE, DEAR, and CGN), in general, no assumption is made about the data except those related to the selected~\gls{dgm}. SCM-VAE and DEAR make the hypothesis that one works with high dimensional data, knowing only a causal structure on some attributes characterizing the data, like labels of images. Both methods hence encode and decode the high-dimensional data into their attribute and model the~\gls{scm} over the attributes. CGN is a computer vision model using a specific~\gls{dag} over images, liking an image shape, texture, and background. Let us also highlight that WhatIfGAN has been designed to deal with an~\gls{scm} with variables of different dimensions (\eg{} scalars and images). Indeed, it trains the generators independently, when not confounded, to facilitate the convergence and enable one to utilize pre-trained models.

\paragraph{Abduction.} On the one hand, as detailed in Section~\ref{sec:taxonomy}, all the explicit (\ie{} invertible explicit and amortized explicit) methods are designed to automatically perform the abduction step. On the other hand, the implicit methods must rely on the sample rejection procedure. However, this abduction procedure is only implemented in NCM. As a result, if one wants to estimate counterfactuals with the other implicit methods, this procedure needs to be implemented on top of the existing code. Note that learning an encoder can replace the sample rejection procedure, though it is unclear how expensive and efficient it is compared to this procedure. Indeed, no comparison has been made yet.

\paragraph{Identifiability guarantees.} Several results on identifiability exist. The most noticeable one is the NCM identifiability theorem, as the NCM class contains all the methods considered in this work except DCM; see Section~\ref{sec:taxonomy}. This result states that an $\mathcal{L}_2$~\citep{xia2022causalneural} or $\mathcal{L}_3$~\citep{xia2023neural} query is identifiable from a $\mathcal{G^{\star}}$-constrained NCM if and only if it is identifiable from the true~\gls{scm}. This can be interpreted as follows: if the search space of the selected feedforward deep conditional generative models is expressive enough to represent the causal mechanisms, the identifiability property depends only on the causal structure and the available data to train the model. Then, a natural question arises: What if the assumptions made about the causal graph are wrong due to imperfect causal knowledge? We shall return to this point in~\Cref{sec:challenges_partial_id}. DCM also provides the same $\mathcal{L}_3$-identifiability result but 
in the absence of hidden confounders. In addition, DCM is the only method to furnish $\mathcal{L}_3$-error bounds based on the encode-decoder reconstruction error after training.

\paragraph{Discussion.} \cref{tb:model_hypotheses} shows that, firstly, half of the method does not implement the abduction step automatically. However, one can rely on the sample rejection procedure~\citep{xia2023neural} to answer $\mathcal{L}_3$ queries. Secondly, only a few methods consider hidden confounders. 
Thirdly, on one hand, in the absence of hidden confounding, knowing a causal ordering and building the associated fully connected~\gls{dag} is sufficient to compute counterfactuals. On the other hand, if there are hidden confounders, it is preferable to precisely locate them in the causal graph and use the NeuralID algorithm of~\cite{xia2023neural} to evaluate if the desired query is identifiable. Fourthly, for all the methods reviewed in this paper, any \(\mathcal{L}_3\)-query is identifiable if it is from the true~\gls{scm} $\mathcal{M}^{\star}$. 
\subsection{Evaluation and Applications}\label{sec:perf}

\begin{table*}[!ht]
    \centering    
    \footnotesize{%
    \begin{tabularx}{\linewidth}{>{\hsize=.16\hsize}C>{\hsize=.24\hsize}X>{\hsize=.09\hsize}C>{\hsize=.17\hsize}C>{\hsize=.34\hsize}X}
      \toprule
        \textbf{Method} & \multicolumn{1}{c}{\textbf{Dataset}} & \multicolumn{1}{c}{\textbf{PCH}} & \multicolumn{1}{c}{\textbf{DSCM Comparison}} & \multicolumn{1}{c}{\textbf{Applications}} \\
      \midrule
      
        \rowcolor{gray!25}
        \textsc{\textit{NF-BGM}} \citetalias{nasr_23} &  
        Ellips generation simulations& 
        $\mathcal{L}_3$ & 
        \xmark &
        Video streaming simulations for adaptive bitrate \\

        \textsc{\textit{NF-DSCM}} \citetalias{pawlowski20} &  
        Morpho-MNIST & 
        $\mathcal{L}_3$ & 
        \xmark &
        Scientific discovery~\citep{yu_23}, cyber-security data generation~\citep{Agrawal2023attack}  \\

        \rowcolor{gray!25}
        \textsc{\textit{GAN-NCM; MLE-NCM}} \citetalias{xia2023neural} &  
        Simulated~\glspl{scm} & 
        $\mathcal{L}_3$ & 
        \checkmark GAN, MLE NCM &
        - \\

        \textsc{\textit{Causal-NF}} \citetalias{javaloy2023causal} &  
        Simulated~\glspl{scm} & 
        $\mathcal{L}_3$ & 
        \checkmark VACA, CARFEL &
        Counterfactual fairness \& fair regularization of classifier\\
        
       \rowcolor{gray!25}
        \textsc{\textit{NCF}} \citetalias{parafita2020causal} & 
        Salary simulations using a simulated~\gls{scm} & 
        $\mathcal{L}_3$ & 
        \xmark &
        Counterfactual fairness and explainability \\

        \textsc{\textit{CARFEL}} \citetalias{khemakhem21a} &  
        4-dimentional polynomial simulated~\gls{scm}, fMRI & 
        $\mathcal{L}_2$ \& $\mathcal{L}_3$ & 
        \xmark &
        - \\
        
        \rowcolor{gray!25}
        \textsc{\textit{iVGAE}} \citetalias{zecevic2021relating} &  
        ASIA & 
        $\mathcal{L}_2$ & 
        \xmark &
        - \\
        
        \textsc{\textit{VACA}} \citetalias{sanchezmartin:22} &  
        Simulated~\glspl{scm}
        & 
        $\mathcal{L}_3$ & 
        \xmark &
        Counterfactual fairness \\

        \rowcolor{gray!25}
        \textsc{\textit{DCM}} \citetalias{chao2023interventional} &  
        Simulated~\glspl{scm}
        , fMRI & 
        $\mathcal{L}_3$ & 
        \checkmark VACA, CARFEL & 
        - \\
        
        \textsc{\textit{SCM-VAE}} \citetalias{Komanduri22} &  
        Pendulum, CelebA & 
        $\mathcal{L}_2$ & 
        \xmark &
        - \\

        \rowcolor{gray!25}
        \textsc{\textit{Causal-TGAN}} \citetalias{wen2022causaltgan} & 
        ASIA, Child, ALARM, Insurance; Adult, Census, News & 
        $\mathcal{L}_1$ & 
        \xmark &
        In-domain data augmentation \\
        
        \textsc{\textit{CausalGAN}} \citetalias{kocaoglu2017causalgan} &  
        CelebA & 
        $\mathcal{L}_2^\star$ & 
        \xmark &
        Out-of-domain data augmentation \\

        \rowcolor{gray!25}
        \textsc{\textit{CFGAN}} \citetalias{ijcai2019p201} &  
        Adult & 
        $\mathcal{L}_2^\sharp$ & 
        \xmark &
        Fairness debiasing \\
        
        \textsc{\textit{DECAF}} \citetalias{vanBreugel21} &  
        Adult, Credit Approval & 
        $\mathcal{L}_2^\sharp$ & 
        \checkmark CFGAN &
        Fairness debiasing \\

        \rowcolor{gray!25}
        \textsc{\textit{WhatIfGAN}} \citetalias{rahman2023towards} &  
        Color-MNIST & 
        $\mathcal{L}_2$ & 
        \checkmark NCM &
        - \\
        
        \textsc{\textit{CGN}} \citetalias{sauer2021counterfactual} &  
        \makecell[l]{Color-MNIST; \\ ImageNet~\citep{imagenet}} & 
        $\mathcal{L}_2^\tau$ & 
        \xmark &
        Out-of-domain data augmentation \\

        \rowcolor{gray!25}
        \textsc{\textit{DEAR}} \citetalias{shen_22} &  
        Pendulum, CelebA & 
        $\mathcal{L}_2^\star$ & 
        \xmark &
        - \\
        
   \cmidrule{1-5}
   \multicolumn{5}{l}{\(^\star\) Disentanglement, \(^\sharp\) Fairness debiasing by intervention, \(^\tau\) Invariant classification after intervention} \\ 
   \bottomrule
 \end{tabularx}
}
\caption{Existing evaluations and applications of~\glspl{dscm}}\label{tb:model_evaluation_nb}
\end{table*}

\paragraph{Experimental evaluation.} There is considerable heterogeneity in the evaluation of the methods: on the datasets, the~\gls{pch} of the tasks, and the metrics. The~\gls{pch} heterogeneity comes from the fact that only half of the methods automatically compute the abduction step. Those methods are hence evaluated on $\mathcal{L}_2$-tasks such as intervention estimation, disentanglement, invariant classification, or debiasing by intervening. fMRI~\citep{Thompson2020}, Color-MNIST~\citep{colored_mnist} are commonly used image datasets to evaluate intervention estimation. Pendulum~\citep{Yang_21} and CelebA~\citep{Liu_2015_ICCV} seem more suited for disentanglement while Morpho-MNIST~\citep{castro_19} can be used for counterfactual estimation. Regarding tabular data, bnlearn~\citep{bnlearn} datasets (ASIA, Child, ALARM, Insurance) are often used to evaluate intervention estimation accuracy as they are extracted from causal bayesian networks. Datasets from the UCI repository~\citep{uci_repo} (Adult, Credit Approval, Census, News) are also popular for real data $\mathcal{L}_2$-experiments as causal graphs can be easily approximated from common sense knowledge. However, regarding $\mathcal{L}_3$-estimations, there is a high heterogeneity between the simulated~\glspl{scm}. As a result, without a unified benchmark, it is a hard task to compare their capabilities to answer counterfactual questions. Moreover, the simulated~\glspl{scm} are designed without multiple sources of randomness~(\eg{}~\gls{dag} structure). This can introduce a bias in the evaluation, resulting in non-generalizable performance when facing real-world applications.

\paragraph{Applications.} Three major goals have been considered in the literature: fairness, explainability, and robustness of machine learning models. Concretely, ImageCFGen\footnote{ImageCFGEN~\citep{Dash_2022_WACV} uses an~\gls{scm} learned with NF-DSCM~\citep{pawlowski20}.}~\citep{Dash_2022_WACV}, Causal-NF, NCF, and VACA were used for counterfactual fairness evaluation. Counterfactual fairness consists of assessing the existence of discrimination regarding a sensitive feature by answering the following counterfactual question: ``Would the outcome have been the same had the person had a different sensitive feature value?''. Once a model is assessed to be counterfactually unfair, counterfactuals can also be used to debiasing a model. For instance, ImageCFGen and Causal-NF have been used to regularize classifiers on unfair classifications by penalizing outcome variation over unfair counterfactuals. Focusing on data fairness instead of model fairness, DECAF and CFGAN have been designed to remove unfair biases from datasets learning the~\gls{scm} modeling the data generation process and generating a new dataset intervening on the causal graph to break unfair causal paths. Regarding explainability, NCF has also been used for counterfactual explanations of black box image classifiers. Causal data augmentation has been a common approach to increase the robustness of machine learning methods for image classification. CausalTGAN performs in-domain augmentation by sampling data from the learned~\gls{scm} without intervention. CausalGAN and CGN rely on out-of-domain data augmentation by generating data from an intervened~\gls{scm}. The interventions are performed on the variables that change the image without changing the label to predict. \cite{cai2023attack} also use out-of-domain data generation for adversarial attacks of image classifiers. \cite{yu_23} use NF-DSCM for scientific discovery in the case of a Bioinformatics application. They use a counterfactual approach to determine which genetic correlations actually correspond to causal effects on Alzheimer and Glioblastoma illnesses. Finally, \cite{Agrawal2023attack} uses~\glspl{dscm} to generate synthetic data to train and test cybersecurity models.

\section{Challenges and Future Directions}\label{sec:challenges}

\subsection{Lack of Evaluation}\label{sec:challenges_eval}

The lack of a benchmark comprises the major challenge one faces when comparing existing~\acrlongpl{dscm}, as underlined in~\cref{tb:model_evaluation_nb}. The methods must be evaluated using the same counterfactual distributions and metrics. In addition, they should be assessed regarding their data efficiency, computing time, and robustness to real-world challenges like selection bias, imbalanced data distributions, and imperfect causal knowledge. Furthermore, assessing the methods' robustness against potential misspecification of the causal graph is fundamental. To support a fair comparative evaluation, the community needs to build a referential~\gls{scm} simulator. Indeed, instead of selecting specific~\glspl{scm}, the simulator will enable one to randomly generate~\glspl{dag}, exogenous noises, and causal mechanisms to fully assess these methods' capabilities, paying attention to the various pitfalls involved in such automatic procedures~\citep{reisach:21}. Extending the evaluation methodology of~\cite{poinsot2023a} to counterfactuals could be a possible direction.\\
\indent Finally, a first step towards such a comprehensible benchmarking of~\glspl{dscm} could focus on the comparison of abduction procedures. Such a study is expected to assess the comparative merit of sample rejection with respect to learning an encoder.
%

\subsection{From Identifiability to Partial Identifiability}\label{sec:challenges_partial_id}

Most of the causal generative methods presented in this survey assume knowing the actual causal graph. Yet, assuming complete knowledge of the causal structure is a strong assumption. Indeed, causal graphs are usually built either based on expert knowledge, which is sensitive to human biases, or using causal discovery algorithms, which still have numerous limitations nowadays~\citep{vowels:22,faller:23}. It is thus essential not to dismiss the possibility of relying on an erroneous causal graph, as this can have profound implications for the validity of causal inference results. Some other assumptions, such as additive noise or absence of selection bias, can also be questioned. An envisioned solution relies on moving towards partial identification, which consists of computing bounds for causal queries whenever point identification is impossible due to unsatisfied or un-testable assumptions. 
Some work has already been done to derive partial identifiability for discrete~\glspl{scm}~\citep{zaffalon20,zhang22}. Hence, more research still needs to be carried out to provide such partial identifiability guarantees to~\glspl{dscm} in general. However, some partial identifiability guarantees have been proven to be non-informative. Indeed, \cite{melnychuk_23} showed, for instance, that, in general, the counterfactual outcome of (un)treated has non-informative bounds in the class of continuous bivariate~\glspl{scm}. Even in these cases, it is still possible to carry out a sensitivity analysis to compute bounds on the causal query. Among the studied methods in this paper, only~\cite{vanBreugel21} carried out a sensitivity analysis regarding the imperfect knowledge of the causal graph. However, the profound impact of any wrong assumption on the validity of causal query estimation should motivate the systematic use of sensitivity analysis for practitioners and theoreticians. 
Note that some work has already been done on sensitivity analysis under hidden confounding. For example, \cite{frauen:23} developed a neural framework for generalized sensitivity analysis under unobserved confounding for $\mathcal{L}_2$-queries including~\gls{cate}. Whereas, \cite{schroder:23} studied the sensitivity of causal fairness ($\mathcal{L}_3$-queries) under unobserved confounding.
%
%
\subsection{Applications: Opportunities and Challenges}

\paragraph{Realistic Causal Simulations.} As done by~\citep{Agrawal2023attack}, we believe~\glspl{dscm} constitute a great opportunity for practitioners to generate synthetic data similar to real ones following a causal generating process. We particularly think that such approaches could be very useful to benchmark causal inference methods on various types of data coming from numerous applications (\eg{} finance, marketing, healthcare), while having a ground truth. Nevertheless, the causal graph, assumed to be known, may, in fact, be misspecified. For this reason, it is crucial to ensure that it was validated by experts beforehand. Otherwise, we recommend limiting the use of such~\glspl{dscm} to realistic synthetic data generation. Moreover, it is important to note that the existing methods have only been designed to deal with tabular data and images.

\paragraph{Sensitive Applications.} As presented in Section~\ref{sec:hyp_guarantees}, there exists an algorithm, NeuralID~\citep{xia2023neural}, compatible with all the methods considered in this survey, automatically testing for query identifiability given the causal graph and the data. This constitutes a great opportunity for sensitive applications requiring identifiability guarantees. However, such an analysis still depends on the hypotheses about the causal graph. This emphasizes again the importance of sensitivity analyses. Moreover, it is also paramount for decision-makers to have access to uncertainty or error measures. DCM is the only method to provide error bounds. Therefore a major direction for forward research is related to uncertainty quantification extending the active research in numerical modeling~\citep{rischel:21,wang:22} to the field of causal inference. In practice, the reconstruction error used in DCM could be extended to amortised explicit models.

\section{Conclusion}\label{sec:conclusion}

This paper reviews and classifies~\acrfullpl{dscm}, examining them from the perspectives of their underlying classes of~\acrfull{scm} and~\acrfull{dgm}. This dual viewpoint enables one to spotlight the guarantees concerning the identifiability and the ability of these models to address counterfactual queries using only observational data and existing causal knowledge. The analyses reveal that the majority of these guarantees are rooted in the intrinsic properties of the deep learning architectures. Moreover, the studied methods have predominantly been evaluated through heterogeneous and straightforward datasets. This highlights the critical need for developing benchmark datasets that encapsulate the complexities encountered in various high-stakes applications, such as in the medical and financial domains. Lastly, our findings underscore the importance of integrating uncertainty quantification methods and partial counterfactual estimation methods with~\gls{dscm}. Such a joint approach could significantly improve the resilience and practical utility of these models, especially in complex environments.
\clearpage
\appendix


\section*{Ethical Statement}\label{sec:ethic}

In this paper, we describe the limitations and guarantees of the discussed~\acrlongpl{dscm}. We strongly advise against using these methods to draw causal conclusions without validation from qualified experts.
\section*{Acknowledgments}\label{sec:akg}
%
This research received partial funding from the European Commission under the Horizon 2020 program (TRUST-AI Project, Contract No. 952060) and (TAILOR, Contract No. 952215), and from the ANR under the France 2030 program, under the reference 23-PEIA-004\@.
%

\bibliographystyle{named}
\bibliography{references_simplified}

\begin{thebibliography}{}

\bibitem[\protect\citeauthoryear{Agrawal \bgroup \em et al.\egroup }{2024}]{Agrawal2023attack}
Garima Agrawal, Amardeep Kaur, and Sowmya Myneni.
\newblock A review of generative models in generating synthetic attack data for cybersecurity.
\newblock {\em Electronics}, 13(2), 2024.

\bibitem[\protect\citeauthoryear{Bareinboim \bgroup \em et al.\egroup }{2022}]{bareinboim:22}
Elias Bareinboim, Juan~D. Correa, Duligur Ibeling, and Thomas Icard.
\newblock {\em {On Pearl's Hierarchy and the Foundations of Causal Inference}}, pages 507--556.
\newblock ACM, 2022.

\bibitem[\protect\citeauthoryear{Cai \bgroup \em et al.\egroup }{2024}]{cai2023attack}
Ruichu Cai, Yuxuan Zhu, Jie Qiao, Zefeng Liang, Furui Liu, and Zhifeng Hao.
\newblock Where and how to attack? a causality-inspired recipe for generating counterfactual adversarial examples.
\newblock In {\em AAAI Conference on Artificial Intelligence}, 2024.

\bibitem[\protect\citeauthoryear{Castro \bgroup \em et al.\egroup }{2019}]{castro_19}
Daniel~C. Castro, Jeremy Tan, Bernhard Kainz, Ender Konukoglu, and Ben Glocker.
\newblock {Morpho-MNIST}: Quantitative assessment and diagnostics for representation learning.
\newblock {\em Journal of Machine Learning Research}, 2019.

\bibitem[\protect\citeauthoryear{Chao \bgroup \em et al.\egroup }{2023}]{chao2023interventional}
Patrick Chao, Patrick Blöbaum, and Shiva~Prasad Kasiviswanathan.
\newblock Interventional and counterfactual inference with diffusion models.
\newblock {\em arxiv:2302.00860}, 2023.
\newblock \textcolor{blue}{Citation alias [9] used in~\cref{tb:model_hypotheses,tb:model_evaluation_nb}}.

\bibitem[\protect\citeauthoryear{Dash \bgroup \em et al.\egroup }{2022}]{Dash_2022_WACV}
Saloni Dash, Vineeth~N Balasubramanian, and Amit Sharma.
\newblock Evaluating and mitigating bias in image classifiers: A causal perspective using counterfactuals.
\newblock In {\em IEEE/CVF Winter Conference on Applications of Computer Vision}, 2022.

\bibitem[\protect\citeauthoryear{Deng \bgroup \em et al.\egroup }{2009}]{imagenet}
Jia Deng, Wei Dong, Richard Socher, Li-Jia Li, Kai Li, and Li~Fei-Fei.
\newblock Imagenet: A large-scale hierarchical image database.
\newblock In {\em IEEE Conference on Computer Vision and Pattern Recognition}, 2009.

\bibitem[\protect\citeauthoryear{Dua and Graff}{2020}]{uci_repo}
Dheeru Dua and Casey Graff.
\newblock {UCI} machine learning repository.
\newblock archive.ics.uci.edu/ml/datasets, 2020.

\bibitem[\protect\citeauthoryear{Faller \bgroup \em et al.\egroup }{2024}]{faller:23}
Philipp~M Faller, Leena~C Vankadara, Atalanti~A Mastakouri, Francesco Locatello, and Dominik Janzing.
\newblock Self-compatibility: Evaluating causal discovery without ground truth.
\newblock In {\em International Conference on Artificial Intelligence and Statistics}, 2024.

\bibitem[\protect\citeauthoryear{Frauen \bgroup \em et al.\egroup }{2024}]{frauen:23}
Dennis Frauen, Fergus Imrie, Alicia Curth, Valentyn Melnychuk, Stefan Feuerriegel, and Mihaela van~der Schaar.
\newblock A neural framework for generalized causal sensitivity analysis.
\newblock In {\em 12th International Conference on Learning Representations}, 2024.

\bibitem[\protect\citeauthoryear{Hornik}{1991}]{HORNIK1991251}
Kurt Hornik.
\newblock Approximation capabilities of multilayer feedforward networks.
\newblock {\em Neural Networks}, 4(2):251--257, 1991.

\bibitem[\protect\citeauthoryear{Ibeling and Icard}{2020}]{ibeling:20}
Duligur Ibeling and Thomas Icard.
\newblock Probabilistic reasoning across the causal hierarchy.
\newblock In {\em AAAI Conference on Artificial Intelligence}, 2020.

\bibitem[\protect\citeauthoryear{Javaloy \bgroup \em et al.\egroup }{2023}]{javaloy2023causal}
Adri{\'a}n Javaloy, Pablo S{\'a}nchez-Mart{\'\i}n, and Isabel Valera.
\newblock Causal normalizing flows: from theory to practice.
\newblock {\em Advances in Neural Information Processing Systems}, 2023.
\newblock \textcolor{blue}{Citation alias [4] used in~\cref{tb:model_hypotheses,tb:model_evaluation_nb}}.

\bibitem[\protect\citeauthoryear{Kaddour \bgroup \em et al.\egroup }{2022}]{kaddour2022causal}
Jean Kaddour, Aengus Lynch, Qi~Liu, Matt~J. Kusner, and Ricardo Silva.
\newblock Causal machine learning: A survey and open problems.
\newblock {\em arXiv:2206.15475}, 2022.

\bibitem[\protect\citeauthoryear{Khemakhem \bgroup \em et al.\egroup }{2021}]{khemakhem21a}
Ilyes Khemakhem, Ricardo Monti, Robert Leech, and Aapo Hyvarinen.
\newblock Causal autoregressive flows.
\newblock In {\em 24th International Conference on Artificial Intelligence and Statistics}, 2021.
\newblock \textcolor{blue}{Citation alias [6] used in~\cref{tb:model_hypotheses,tb:model_evaluation_nb}}.

\bibitem[\protect\citeauthoryear{Kim \bgroup \em et al.\egroup }{2019}]{colored_mnist}
Byungju Kim, Hyunwoo Kim, Kyungsu Kim, Sungjin Kim, and Junmo Kim.
\newblock Learning not to learn: Training deep neural networks with biased data.
\newblock In {\em IEEE/CVF Conference on Computer Vision and Pattern Recognition}, 2019.

\bibitem[\protect\citeauthoryear{Kocaoglu \bgroup \em et al.\egroup }{2018}]{kocaoglu2017causalgan}
Murat Kocaoglu, Christopher Snyder, Alexandros~G. Dimakis, and Sriram Vishwanath.
\newblock Causal{GAN}: Learning causal implicit generative models with adversarial training.
\newblock In {\em International Conference on Learning Representations}, 2018.
\newblock \textcolor{blue}{Citation alias [12] used in~\cref{tb:model_hypotheses,tb:model_evaluation_nb}}.

\bibitem[\protect\citeauthoryear{Komanduri \bgroup \em et al.\egroup }{2022}]{Komanduri22}
Aneesh Komanduri, Yongkai Wu, Wen Huang, Feng Chen, and Xintao Wu.
\newblock {SCM-VAE}: Learning identifiable causal representations via structural knowledge.
\newblock In {\em IEEE International Conference on Big Data}, 2022.
\newblock \textcolor{blue}{Citation alias [10] used in~\cref{tb:model_hypotheses,tb:model_evaluation_nb}}.

\bibitem[\protect\citeauthoryear{Komanduri \bgroup \em et al.\egroup }{2023}]{komanduri:survey:23}
Aneesh Komanduri, Xintao Wu, Yongkai Wu, and Feng Chen.
\newblock From identifiable causal representations to controllable counterfactual generation: A survey on causal generative modeling.
\newblock {\em arXiv:2310.11011}, 2023.

\bibitem[\protect\citeauthoryear{Liu \bgroup \em et al.\egroup }{2015}]{Liu_2015_ICCV}
Ziwei Liu, Ping Luo, Xiaogang Wang, and Xiaoou Tang.
\newblock Deep learning face attributes in the wild.
\newblock In {\em IEEE International Conference on Computer Vision}, 2015.

\bibitem[\protect\citeauthoryear{Melnychuk \bgroup \em et al.\egroup }{2023}]{melnychuk_23}
Valentyn Melnychuk, Dennis Frauen, and Stefan Feuerriegel.
\newblock Partial counterfactual identification of continuous outcomes with a curvature sensitivity model.
\newblock In {\em 36 Advances in Neural Information Processing Systems}, 2023.

\bibitem[\protect\citeauthoryear{Nasr-Esfahany \bgroup \em et al.\egroup }{2023}]{nasr_23}
Arash Nasr-Esfahany, Mohammad Alizadeh, and Devavrat Shah.
\newblock Counterfactual identifiability of bijective causal models.
\newblock In {\em 40th International Conference on Machine Learning}, 2023.
\newblock \textcolor{blue}{Citation alias [1] used in~\cref{tb:model_hypotheses,tb:model_evaluation_nb}}.

\bibitem[\protect\citeauthoryear{Parafita and Vitri{\`a}}{2020}]{parafita2020causal}
{\'A}lvaro Parafita and Jordi Vitri{\`a}.
\newblock Causal inference with deep causal graphs.
\newblock {\em arXiv:2006.08380}, 2020.
\newblock \textcolor{blue}{Citation alias [5] used in~\cref{tb:model_hypotheses,tb:model_evaluation_nb}}.

\bibitem[\protect\citeauthoryear{Pawlowski \bgroup \em et al.\egroup }{2020}]{pawlowski20}
Nick Pawlowski, Daniel Coelho~de Castro, and Ben Glocker.
\newblock Deep structural causal models for tractable counterfactual inference.
\newblock In {\em Advances in Neural Information Processing Systems}, 2020.
\newblock \textcolor{blue}{Citation alias [2] used in~\cref{tb:model_hypotheses,tb:model_evaluation_nb}}.

\bibitem[\protect\citeauthoryear{Pearl and Mackenzie}{2018}]{pearl:18}
Judea Pearl and Dana Mackenzie.
\newblock {\em The book of why: the new science of cause and effect}.
\newblock Basic books, 2018.

\bibitem[\protect\citeauthoryear{Pearl}{1995}]{pearl:95}
Judea Pearl.
\newblock Causal diagrams for empirical research.
\newblock {\em Biometrika}, 1995.

\bibitem[\protect\citeauthoryear{Pearl}{2009a}]{pearl:09survey}
Judea Pearl.
\newblock Causal inference in statistics: an overview.
\newblock {\em Statistics Surveys}, 2009.

\bibitem[\protect\citeauthoryear{Pearl}{2009b}]{pearl:09}
Judea Pearl.
\newblock {\em Causality: Models, Reasoning and Inference}.
\newblock Cambridge University Press, 2nd edition, 2009.

\bibitem[\protect\citeauthoryear{Poinsot and Leite}{2023}]{poinsot2023a}
Audrey Poinsot and Alessandro Leite.
\newblock A guide for practical use of {ADMG} causal data augmentation.
\newblock In {\em ICLR 2023 Workshop on Pitfalls of limited data and computation for Trustworthy ML}, 2023.

\bibitem[\protect\citeauthoryear{Rahman and Kocaoglu}{2023}]{rahman2023towards}
Md~Musfiqur Rahman and Murat Kocaoglu.
\newblock Towards modular learning of deep causal generative models.
\newblock In {\em ICML Workshop on Structured Probabilistic Inference {\&} Generative Modeling}, 2023.
\newblock \textcolor{blue}{Citation alias [15] used in~\cref{tb:model_hypotheses,tb:model_evaluation_nb}}.

\bibitem[\protect\citeauthoryear{Reisach \bgroup \em et al.\egroup }{2021}]{reisach:21}
Alexander~G. Reisach, Christof Seiler, and Sebastian Weichwald.
\newblock {Beware of the Simulated DAG! Causal Discovery Benchmarks May Be Easy to Game}.
\newblock In {\em {34th Advances in Neural Information Processing Systems}}, 2021.

\bibitem[\protect\citeauthoryear{Rischel and Weichwald}{2021}]{rischel:21}
Eigil~F. Rischel and Sebastian Weichwald.
\newblock Compositional abstraction error and a category of causal models.
\newblock In {\em 37th Conference on Uncertainty in Artificial Intelligence}, volume 161, 2021.

\bibitem[\protect\citeauthoryear{S{\'a}nchez-Martin \bgroup \em et al.\egroup }{2022}]{sanchezmartin:22}
Pablo S{\'a}nchez-Martin, Miriam Rateike, and Isabel Valera.
\newblock {VACA}: designing variational graph autoencoders for causal queries.
\newblock In {\em AAAI Conference on Artificial Intelligence}, 2022.
\newblock \textcolor{blue}{Citation alias [8] used in~\cref{tb:model_hypotheses,tb:model_evaluation_nb}}.

\bibitem[\protect\citeauthoryear{Sauer and Geiger}{2021}]{sauer2021counterfactual}
Axel Sauer and Andreas Geiger.
\newblock Counterfactual generative networks.
\newblock In {\em International Conference on Learning Representations}, 2021.
\newblock \textcolor{blue}{Citation alias [16] used in~\cref{tb:model_hypotheses,tb:model_evaluation_nb}}.

\bibitem[\protect\citeauthoryear{Schr{\"o}der \bgroup \em et al.\egroup }{2023}]{schroder:23}
Maresa Schr{\"o}der, Dennis Frauen, and Stefan Feuerriegel.
\newblock Causal fairness under unobserved confounding: a neural sensitivity framework.
\newblock {\em 12th International Conference on Learning Representations}, 2023.

\bibitem[\protect\citeauthoryear{Scutari}{2010}]{bnlearn}
Marco Scutari.
\newblock Learning bayesian networks with the bnlearn {R} package.
\newblock In {\em Journal of Statistical Software}, 2010.

\bibitem[\protect\citeauthoryear{Shen \bgroup \em et al.\egroup }{2022}]{shen_22}
Xinwei Shen, Furui Liu, Hanze Dong, Qing Lian, Zhitang Chen, and Tong Zhang.
\newblock Weakly supervised disentangled generative causal representation learning.
\newblock {\em Journal of Machine Learning Research}, 2022.
\newblock \textcolor{blue}{Citation alias [17] used in~\cref{tb:model_hypotheses,tb:model_evaluation_nb}}.

\bibitem[\protect\citeauthoryear{Sohl-Dickstein \bgroup \em et al.\egroup }{2015}]{sohl:15deep}
Jascha Sohl-Dickstein, Eric Weiss, Niru Maheswaranathan, and Surya Ganguli.
\newblock Deep unsupervised learning using nonequilibrium thermodynamics.
\newblock In {\em International Conference on Machine Learning}, 2015.

\bibitem[\protect\citeauthoryear{Spirtes \bgroup \em et al.\egroup }{2000}]{spirtes:00}
Peter Spirtes, Clark~N Glymour, and Richard Scheines.
\newblock {\em Causation, prediction, and search}.
\newblock The MIT Press, 2000.

\bibitem[\protect\citeauthoryear{Thompson \bgroup \em et al.\egroup }{2020}]{Thompson2020}
WH~Thompson, R~Nair, H~Oya, O~Esteban, JM~Shine, CI~Petkov, RA~Poldrack, M~Howard, and R~Adolphs.
\newblock Human {es-fMRI} resource: Concurrent deep-brain stimulation and whole-brain functional {MRI}.
\newblock {\em bioRxiv}, 2020.

\bibitem[\protect\citeauthoryear{van Breugel \bgroup \em et al.\egroup }{2021}]{vanBreugel21}
Boris van Breugel, Trent Kyono, Jeroen Berrevoets, and Mihaela van~der Schaar.
\newblock {DECAF}: Generating fair synthetic data using causally-aware generative networks.
\newblock In {\em Advances in Neural Information Processing Systems}, 2021.
\newblock \textcolor{blue}{Citation alias [14] used in~\cref{tb:model_hypotheses,tb:model_evaluation_nb}}.

\bibitem[\protect\citeauthoryear{Vowels \bgroup \em et al.\egroup }{2022}]{vowels:22}
Matthew~J. Vowels, Necati~Cihan Camgoz, and Richard Bowden.
\newblock D'ya like {DAGs}? a survey on structure learning and causal discovery.
\newblock {\em ACM Computing Surveys}, 2022.

\bibitem[\protect\citeauthoryear{Wang \bgroup \em et al.\egroup }{2022}]{wang:22}
Benjie Wang, Matthew~R Wicker, and Marta Kwiatkowska.
\newblock Tractable uncertainty for structure learning.
\newblock In {\em 39th International Conference on Machine Learning}, 2022.

\bibitem[\protect\citeauthoryear{Wen \bgroup \em et al.\egroup }{2022}]{wen2022causaltgan}
Bingyang Wen, Yupeng Cao, Fan Yang, Koduvayur Subbalakshmi, and Rajarathnam Chandramouli.
\newblock Causal-{TGAN}: Modeling tabular data using causally-aware {GAN}.
\newblock In {\em ICLR Workshop on Deep Generative Models for Highly Structured Data}, 2022.
\newblock \textcolor{blue}{Citation alias [11] used in~\cref{tb:model_hypotheses,tb:model_evaluation_nb}}.

\bibitem[\protect\citeauthoryear{Xia \bgroup \em et al.\egroup }{2021}]{xia2022causalneural}
Kevin Xia, Kai-Zhan Lee, Yoshua Bengio, and Elias Bareinboim.
\newblock The causal-neural connection: Expressiveness, learnability, and inference.
\newblock In {\em Advances in Neural Information Processing Systems}, 2021.

\bibitem[\protect\citeauthoryear{Xia \bgroup \em et al.\egroup }{2023}]{xia2023neural}
Kevin~Muyuan Xia, Yushu Pan, and Elias Bareinboim.
\newblock Neural causal models for counterfactual identification and estimation.
\newblock In {\em 11th International Conference on Learning Representations}, 2023.
\newblock \textcolor{blue}{Citation alias [3] used in~\cref{tb:model_hypotheses,tb:model_evaluation_nb}}.

\bibitem[\protect\citeauthoryear{Xu \bgroup \em et al.\egroup }{2019}]{ijcai2019p201}
Depeng Xu, Yongkai Wu, Shuhan Yuan, Lu~Zhang, and Xintao Wu.
\newblock Achieving causal fairness through generative adversarial networks.
\newblock In {\em 28th International Joint Conference on Artificial Intelligence}, 2019.
\newblock \textcolor{blue}{Citation alias [13] used in~\cref{tb:model_hypotheses,tb:model_evaluation_nb}}.

\bibitem[\protect\citeauthoryear{Yang \bgroup \em et al.\egroup }{2021}]{Yang_21}
Mengyue Yang, Furui Liu, Zhitang Chen, Xinwei Shen, Jianye Hao, and Jun Wang.
\newblock {CausalVAE}: Disentangled representation learning via neural structural causal models.
\newblock In {\em IEEE/CVF Conference on Computer Vision and Pattern Recognition}, 2021.

\bibitem[\protect\citeauthoryear{Yu \bgroup \em et al.\egroup }{2023}]{yu_23}
Fanyang Yu, Rongguang Wang, Pratik Chaudhari, and Christos Davatzikos.
\newblock Investigating causality between genotype and clinical phenotype in neurological disorders using structural causal model and normalizing flow.
\newblock In {\em 1st Workshop on Deep Generative Models for Health}, 2023.

\bibitem[\protect\citeauthoryear{Zaffalon \bgroup \em et al.\egroup }{2020}]{zaffalon20}
Marco Zaffalon, Alessandro Antonucci, and Rafael Caba\~nas.
\newblock Structural causal models are (solvable by) credal networks.
\newblock In {\em 10th International Conference on Probabilistic Graphical Models}, 2020.

\bibitem[\protect\citeauthoryear{Zečević \bgroup \em et al.\egroup }{2021}]{zecevic2021relating}
Matej Zečević, Devendra~Singh Dhami, Petar Veličković, and Kristian Kersting.
\newblock Relating graph neural networks to structural causal models.
\newblock {\em arXiv:2109.04173}, 2021.
\newblock \textcolor{blue}{Citation alias [7] used in~\cref{tb:model_hypotheses,tb:model_evaluation_nb}}.

\bibitem[\protect\citeauthoryear{Zhang \bgroup \em et al.\egroup }{2022}]{zhang22}
Junzhe Zhang, Jin Tian, and Elias Bareinboim.
\newblock Partial counterfactual identification from observational and experimental data.
\newblock In {\em 39th International Conference on Machine Learning}, 2022.

\bibitem[\protect\citeauthoryear{Zhou \bgroup \em et al.\egroup }{2023}]{zhou2023emerging}
Guanglin Zhou, Shaoan Xie, Guangyuan Hao, Shiming Chen, Biwei Huang, Xiwei Xu, Chen Wang, Liming Zhu, Lina Yao, and Kun Zhang.
\newblock Emerging synergies in causality and deep generative models: A survey.
\newblock {\em arXiv:2301.12351}, 2023.

\end{thebibliography}

\end{document}